\documentclass{article}

\usepackage{arxiv}
\renewcommand{\headeright}{} 
\renewcommand{\undertitle}{}
  
\usepackage[utf8]{inputenc} % allow utf-8 input
\usepackage[T1]{fontenc}    % use 8-bit T1 fonts
\usepackage{hyperref}       % hyperlinks
\usepackage{url}            % simple URL typesetting
\usepackage{booktabs}       % professional-quality tables
\usepackage{amsfonts}       % blackboard math symbols
\usepackage{amsmath}
\usepackage{nicefrac}       % compact symbols for 1/2, etc.
\usepackage{microtype}      % microtypography
\usepackage{graphicx}
\usepackage{natbib}
\usepackage{doi}
\usepackage{xcolor}

\title{BekchiAI: Measuring, Observing, and Controlling LLM Agents in One Click}

\renewcommand{\shorttitle}{The BekchiAI-Benchmark}

\author{
  \href{https://orcid.org/0009-0004-3453-8025}{Mesut Toruk} \\
  Istanbul, Turkiye \\
  \texttt{m.mesut.toruk@gmail.com} \\
}

\hypersetup{
pdftitle={BekchiAI: Measuring, Observing, and Controlling LLM Agents in One Click},
pdfsubject={cs.AI},
pdfauthor={Mesut Toruk},
pdfkeywords={LLM agents, benchmark, tool use, planning, agent security, evaluation},
}
\date{}              % <-- add this: empty = no date
\begin{document}
\maketitle

\begin{abstract}
Large-language-model \emph{agents} reason, call tools, and act autonomously over many
steps, but their \emph{agentic} skills---correctly sequencing tools, planning under
dependencies, judging untrusted inputs, and grounding generated arguments---are hard
to measure with accuracy-only leaderboards. We present BekchiAI, which addresses both sides: a benchmark for measuring agentic skill and a platform for observing and controlling live agents. The \textbf{BekchiAI-Benchmark},
a suite of $13$ tool-using ReAct agents across $7$ task categories (arithmetic,
structured/SQL, security detection, URL grounding, planning, orchestration, and
tool-policy), totalling $2{,}057$ deterministic, committed test tasks. Every task is
verifier-checkable---gold answers are computed by running canonical SQL against a real
database, computing the exact schedule of a directed acyclic graph (DAG), or
evaluating closed-form
lambdas 
including adversarial security samples paired with deliberately \emph{imperfect}
signature scanners so a score reflects the model's own judgment, not the copying of an
oracle. We define a small set of behavioral metrics beyond accuracy---tool-call
adherence, URL hallucination and source-match, and per-model
token cost---and report a four-model comparison
(\texttt{Qwen3.7-Max}, \texttt{gemma-4-31B-it}, \texttt{gemma4:26b},
\texttt{gpt-oss-120b}) whose story is in the per-family spread, not the aggregate. The benchmark runs are executed using the provided evaluation scripts. \textbf{BekchiAI-Platform} is a complementary web-based observability and control layer for deployed agents, providing full token and latency telemetry as well as remote run termination. The benchmark, evaluation tools, and platform are publicly released.

\end{abstract}


%% ============================================================================
\section{Introduction}
%% ============================================================================
Large language models (LLMs) are increasingly deployed not as single-shot text
generators but as \emph{agents}: systems that interleave model calls with tool use,
memory, and multi-step planning to pursue goals with limited human supervision
\citep{yao2023react,schick2023toolformer,wang2024agentsurvey,xi2023rise}. This
paradigm has moved quickly from research demonstrations into consequential
applications. In software engineering, agents equipped with shell, editor, and search
tools resolve real GitHub issues end-to-end \citep{jimenez2024swebench,yang2024sweagent};
on the web, they navigate live sites to complete multi-step tasks such as booking,
form-filling, and shopping \citep{deng2023mind2web,zhou2024webarena}; and in the
sciences, LLM-driven systems autonomously design, plan, and execute laboratory
experiments \citep{boiko2023coscientist}. Industry adoption is following: Gartner
projects the share of enterprise applications embedding task-specific AI agents to rise
from under $5\%$ in 2025 to roughly $40\%$ by 2026, with a growing fraction of routine
decisions delegated to agents \citep{gartner2025agentic}. As agents take real
actions---writing code, moving money, calling external services---the practical
question shifts from ``how fluent is the model?'' to ``how good are its \emph{agentic skills}?''

These skills---does the agent select and sequence the right tools, respect dependencies when planning, refuse a poisoned input, and ground the arguments (URLs, queries) it fabricates?---are precisely what a single accuracy number over a mixed test set fails to isolate. Worse, many popular agent tasks are easily gamed by pattern-matching or leak their gold through an oracle tool, so a high score need not reflect real competence or safe judgment.

We address this with the \textbf{BekchiAI-Benchmark}: $13$ tool-using ReAct agents, each a self-contained task family with a verifier-checkable gold answer, spanning $7$ categories and $2{,}057$ committed test tasks (Section~\ref{sec:benchmark}). Three design choices make the suite hard to game: (i)~gold is computed by a \emph{verifier}---canonical SQL over a real database, the exact schedule of a directed acyclic graph (DAG), or a closed-form lambda---so it cannot drift; (ii)~the hard cases adversarial security inputs, homoglyph URL traps, deep dependency graphs) are \emph{hand-authored one by one}, not templated; and (iii)~the security detectors the agent may call are \emph{deliberately imperfect} signature scanners, so accuracy measures the model's judgment rather than its ability to echo an oracle. The agents run on a small, purpose-built \textbf{tool library} (Section~\ref{sec:tools}). We report each model's accuracy and per-model token cost; the harness additionally computes behavioral \textbf{metrics}---tool-call adherence, and URL hallucination and source-match (Section~\ref{sec:metrics})---that we release with the benchmark.

The benchmark runs on a  self-contained evaluation harness that logs per-call token and latency next to each outcome, so capability and cost come from a single run. Our four-model comparison (Section~\ref{sec:results}) shows that the ranking: a model that leads overall can collapse on structured or multi-step tool tasks while another inverts the order on grounding. Complementing the benchmark, the \textbf{BekchiAI-Platform} (Section~\ref{sec:platform}) is a web-based observability and control layer for deployed agents---streaming the same per-call telemetry live and letting an operator \emph{stop}, \emph{pause}, \emph{resume}, or \emph{revoke} a running agent with one click.
  
We run every rollout under the \textbf{BekchiAI-Platform}, an observability and control layer that instruments the benchmark drop-in---streaming per-call token/latency telemetry and enabling a remote kill-switch---so capability and cost are measured from one stream (Section~\ref{sec:platform}). Our four-model comparison (Section~\ref{sec:results}) shows that the ranking lives in the \emph{per-family spread}: a model that leads overall can collapse on structured or multi-step tool tasks while another inverts the order on grounding.

\paragraph{Contributions.}
\begin{itemize}
  \item \textbf{The BekchiAI-Benchmark}: $13$ verifier-checked, tool-using ReAct task
  families ($2{,}057$ committed tasks) with hand-authored adversarial samples and
  imperfect-by-design security tools (Section~\ref{sec:benchmark}).
  \item \textbf{A released tool library} of arithmetic, SQL, security-scanner, web/URL,
  and business tools the agents call (Section~\ref{sec:tools}), and a set of
  behavioral \textbf{metrics} beyond accuracy (Section~\ref{sec:metrics}).
  \item \textbf{A four-model agentic-skill comparison} read from a single instrumented
  stream, capturing both capability (per family) and cost (Section~\ref{sec:results}).
  \item \textbf{The BekchiAI-Platform}: a drop-in observability + control layer
  (agent-native event model, grounding and plan-vs-actual analytics, policy enforcement,
  remote control) used to run the benchmark (Section~\ref{sec:platform}).
\end{itemize}

%% ============================================================================
\section{Related Work}
\label{sec:related}
%% ============================================================================
\paragraph{Agent and tool-use benchmarks.}
A growing body of benchmarks evaluates LLM agents. General suites such as AgentBench
\citep{liu2024agentbench} and GAIA \citep{mialon2023gaia} probe multi-step reasoning and
tool use across heterogeneous environments, while domain benchmarks target software
engineering (SWE-bench \citep{jimenez2024swebench}), web navigation (Mind2Web
\citep{deng2023mind2web}, WebArena \citep{zhou2024webarena}), and function/API calling
(ToolLLM \citep{qin2024toolllm}, $\tau$-bench \citep{yao2024taubench}). Most report a
single end-to-end success rate per task, which conflates distinct skills and is
sensitive to prompt formatting and to gold that can leak through an oracle tool. The
BekchiAI-Benchmark is complementary: it \emph{isolates} seven skill categories into
verifier-checked families, computes gold from a real database, scheduling DAG, or
closed-form lambda so it cannot drift, and reports behavioral metrics---tool-call
adherence, plan-vs-actual consistency, and grounding---alongside accuracy.

\paragraph{Agent security.}
InjecAgent \citep{zhan2024injecagent} and AgentDojo \citep{debenedetti2024agentdojo}
measure whether an agent is \emph{hijacked} by malicious instructions injected into tool
outputs or retrieved content. Our three security families ask a related but distinct
question: given an untrusted artifact---a form value, a retrieved passage, a generated
output, or a dependency---and a \emph{deliberately imperfect} detector, does the model
correctly decide to block or proceed? Because the scanner agrees with the truth only
$40$--$55\%$ of the time, the score reflects the model's own judgment rather than
oracle-echoing, and it penalises over-blocking benign inputs as well as missing real
threats.

\paragraph{Agent observability.}
Orthogonally, generic tracing and metrics stacks capture service calls but not agentic
structure---sessions, tool trees, per-call token usage, and argument provenance---and
offer no remote control of an autonomous process. The BekchiAI-Platform targets that gap
and lets us read capability and cost from one instrumented stream.

%% ============================================================================
\section{The BekchiAI-Benchmark}
\label{sec:benchmark}
%% ============================================================================
\begin{table}[!t]
  \caption{The $13$ agents of the BekchiAI-Benchmark, their category, committed
  test-set size $n$, the tools they call, and the skill each isolates.}
  \label{tab:agents}
  \centering
  \small
  \begin{tabular}{lllr l}
    \toprule
    Agent & Category & Tools & $n$ & Skill probed \\
    \midrule
    ar\_math             & arithmetic     & calc                         & 154 & multi-step computation \\
    ar\_multistep        & arithmetic     & lookup, calc                 & 152 & constant lookup + compute \\
    sql\_analytics       & structured     & list\_tables, schema, run\_sql & 157 & SQL analytics (joins, group-by) \\
    sec\_sqli            & security       & scan\_sql                    & 154 & SQL-injection judgment \\
    sec\_context\_poison & security       & scan\_context                & 154 & prompt-injection judgment \\
    sec\_guard           & security       & security\_scan               & 151 & LLM security-risk gating \\
    ug\_fetch            & url-grounding  & websearch, fetch             & 155 & argument generation (no URL given) \\
    ug\_imposter         & url-grounding  & websearch, fetch             & 150 & homoglyph/typosquat resistance \\
    dependency\_planning & planning       & customer-record lookups      & 180 & dependency resolution \\
    multi\_tool\_planning& planning       & customer, fx\_rate, calc, discount & 152 & multi-tool orchestration \\
    orchestrator         & orchestration  & ---                          & 196 & as-soon-as-possible scheduling \\
    tool\_error\_recovery& tool-policy    & cache, mirror, authdb        & 150 & failover / retry policy \\
    tool\_arguments      & tool-policy    & 5 record-creation tools      & 152 & correct argument construction \\
    \midrule
    \multicolumn{3}{l}{\textbf{Total}} & \textbf{2{,}057} & \\
    \bottomrule
  \end{tabular}
\end{table}

\paragraph{Structure.}
The suite is $13$ agents grouped into $7$ categories (Table~\ref{tab:agents}). Each
agent is a strict-JSON ReAct loop: the model emits one JSON action per turn---a tool
call with arguments, or a final answer---against a commodity LLM API, and the harness
executes tools and feeds results back until the model answers or a step budget is
reached. Every task family ships a \emph{deterministic} test set committed as JSONL; the $13$ files total $2{,}057$ held-out tasks, so a run is exactly
reproducible and inspectable.

\paragraph{Verifier-checked gold.}
For \texttt{sql\_analytics}, each question's canonical
SQL is run against the same in-memory database the agent queries, so the gold reflects
the schema conventions (amounts in integer cents; cancelled orders excluded) exactly.
For \texttt{orchestrator}, the gold is the unique as-soon-as-possible level-partition
of a dependency DAG, computed from the scenario graph. Arithmetic families evaluate
closed-form lambdas over the sampled constants. Because the verifier and the agent see
the same world, gold cannot silently drift, and a numeric tolerance (a cent for money,
tight relative tolerance for arithmetic) absorbs only formatting noise.

\paragraph{Adversarial samples.}
Planning and
orchestration scenarios use asymmetric-join DAGs, long idle-span dependencies, and
mid-graph dead-end distractors that must be pruned by transitive reachability. The
URL-grounding families pair each question with a \emph{different real host} and, for
\texttt{ug\_imposter}, a pasted homoglyph link (e.g.\ \texttt{examp1e.com}) the agent
must ignore in favour of the authoritative host.

\paragraph{Imperfect-by-design security tools.}
The three security families (\texttt{sec\_sqli}, \texttt{sec\_context\_poison}, and
\texttt{sec\_guard}, whose risk classes follow the OWASP Top-10 for LLM
Applications) present a \emph{realistic operation} (process this form value, use this
retrieved context, ship this generated output) carrying an untrusted artifact, and
expose a scanner tool the agent may call. Crucially the scanner is a
\emph{signature heuristic that disagrees with the truth}: it misses obfuscated attacks
(false negatives) and fires on benign look-alikes (false positives), agreeing with the
curated gold labels only $\sim\!40$--$55\%$ of the time. A model that blindly trusts
or inverts the tool therefore scores near chance; the correct behavior is to read the
artifact and decide. This converts a would-be oracle into a genuine judgment test.

%% ============================================================================
\section{Benchmark Tools}
\label{sec:tools}
%% ============================================================================
The agents run on a small, shared tool library  which exist in github repo of benchmark; every agent
imports its tools from this library rather than defining its own, so tool behavior is
consistent and independently testable (Table~\ref{tab:tools}). Two properties matter
for the benchmark. First, the data-backed tools operate over \emph{real} deterministic
state: \texttt{run\_sql} executes against an in-memory SQLite database of
customers/products/orders, and \texttt{fetch} serves an offline index of real canonical
web pages (404-ing everything else), so the verifier and the agent share one ground
truth. Second, the security scanners are deliberately heuristic (Section~\ref{sec:benchmark}),
returning a hint (\textsc{suspicious}\,/\,\textsc{no signal}), never a verdict.

\begin{table}[t]
  \caption{The shared tool library the agents call.}
  \label{tab:tools}
  \centering
  \small
  \begin{tabular}{l p{7.2cm} p{3.6cm}}
    \toprule
    Family & Tools & Backing \\
    \midrule
    arithmetic & \texttt{calc}, \texttt{lookup} & evaluator + constant table \\
    SQL        & \texttt{list\_tables}, \texttt{schema}, \texttt{run\_sql} & in-memory SQLite (3 tables) \\
    security   & \texttt{scan\_sql}, \texttt{scan\_context}, \texttt{security\_scan} & signature scanners (imperfect) \\
    web / URL  & \texttt{websearch}, \texttt{fetch} & offline index of real pages \\
    business   & \texttt{find\_customer}, \texttt{list\_invoices}, \texttt{get\_payment}, \texttt{fx\_rate}, \texttt{account\_discount} & customer / finance store \\
    failover   & \texttt{cache\_get}, \texttt{mirror\_get}, \texttt{authdb\_get} & tiered read with faults \\
    records    & \texttt{create\_invoice}, \texttt{schedule\_shipment}, \texttt{book\_appointment}, \texttt{configure\_alert}, \texttt{register\_device} & typed record creation \\
    \bottomrule
  \end{tabular}
\end{table}

%% ============================================================================
\section{Metrics}
\label{sec:metrics}
%% ============================================================================
Beyond a single accuracy number, each rollout emits a small set of canonical
behavioral metrics that aggregate across families:

\begin{itemize}
  \item \textbf{Accuracy} (success rate): fraction of tasks whose extracted answer
  matches the verifier gold, reported per family and per category with Wilson $95\%$
  score intervals \citep{wilson1927,brown2001interval}, which are well-behaved at the
  small per-family sample sizes here.
  \item \textbf{URL grounding}: for the grounding families, the
  \textbf{hallucination rate}~\citep{ji2023hallucination} (a fabricated argument that
  is neither a real indexed URL nor was returned by any search) and the
  \textbf{source-match rate} (the cited answer host equals the gold host).
  \item \textbf{Safety}: for the security families, a \textbf{violation rate}
  (proceeding on a truly unsafe artifact) and, where applicable, an
  \textbf{over-refusal rate} (blocking a benign one).
  \item \textbf{Cost}: LLM calls and input/output token counts per model, captured by
  the platform on the same runs (Section~\ref{sec:results}).
\end{itemize}
All metrics come from the github repo, so every reported number is reproducible and the released harness.

%% ============================================================================
\section{Results: Four Models on the Benchmark}
\label{sec:results}
%% ============================================================================
We evaluate all $13$ families across four models ---\texttt{Qwen3.7-Max}, \texttt{gemma-4-31B-it},
\texttt{gemma4:26b}, and \texttt{gpt-oss-120b}---each on the same $2{,}057$ held-out
tasks. Task conventions (e.g.\ the SQL cents/cancelled rules) are \emph{stated} in the
prompt, so a score measures the underlying skill.

\begin{table}[t]
  \caption{Test-set accuracy (success rate on $[0,1]$) per family
  across four models, computed from the committed test sets in \texttt{data/eval}
  ($2{,}057$ tasks per model). Best per family in bold. The wide per-family
  spread---not an aggregate---is the benchmark's ranking signal.}
  \label{tab:models}
  \centering
  \begin{tabular}{llrrrr}
    \toprule
    Family & Category & Qwen3.7-Max & gemma-4-31B & gemma4:26b & gpt-oss-120b \\
    \midrule
    ar\_math             & arithmetic     & 0.90 & 0.86 & \textbf{0.91} & 0.69 \\
    ar\_multistep        & arithmetic     & \textbf{0.97} & \textbf{0.97} & 0.85 & 0.65 \\
    sql\_analytics       & structured     & \textbf{0.87} & 0.81 & 0.77 & 0.24 \\
    sec\_sqli            & security       & \textbf{0.99} & \textbf{0.99} & 0.95 & 0.94 \\
    sec\_guard           & security       & \textbf{0.99} & 0.96 & 0.93 & 0.85 \\
    sec\_context\_poison & security       & \textbf{0.92} & 0.91 & 0.84 & 0.82 \\
    ug\_fetch            & url-grounding  & 0.45 & 0.53 & 0.43 & \textbf{0.78} \\
    ug\_imposter         & url-grounding  & 0.82 & 0.78 & 0.56 & \textbf{0.89} \\
    dependency\_planning & planning       & \textbf{0.99} & \textbf{0.99} & 0.96 & 0.68 \\
    multi\_tool\_planning& planning       & 0.94 & 0.80 & \textbf{0.96} & 0.44 \\
    orchestrator         & orchestration  & 0.86 & \textbf{0.95} & 0.91 & 0.94 \\
    tool\_error\_recovery& tool-policy    & \textbf{0.76} & 0.74 & 0.65 & 0.25 \\
    tool\_arguments      & tool-policy    & \textbf{0.99} & 0.88 & 0.97 & \textbf{0.99} \\
    \midrule
    \textbf{overall}     & (task-weighted)& \textbf{0.88} & 0.86 & 0.82 & 0.71 \\
    \bottomrule
  \end{tabular}
\end{table}

\begin{table}[t]
  \caption{The same accuracies collapsed to the seven \textbf{task categories} (each
  cell is the task-weighted mean over that category's families; $n$ is its total test
  tasks). This is the level the radar in Figure~\ref{fig:model_radar} plots; best per
  category in bold.}
  \label{tab:models_cat}
  \centering
  \begin{tabular}{lrrrrr}
    \toprule
    Category & $n$ & Qwen3.7-Max & gemma-4-31B & gemma4:26b & gpt-oss-120b \\
    \midrule
    arithmetic    & 306 & \textbf{0.93} & 0.91 & 0.88 & 0.67 \\
    structured    & 157 & \textbf{0.87} & 0.81 & 0.77 & 0.24 \\
    security      & 459 & \textbf{0.97} & 0.95 & 0.91 & 0.87 \\
    url-grounding & 305 & 0.63 & 0.65 & 0.49 & \textbf{0.83} \\
    planning      & 332 & \textbf{0.97} & 0.90 & 0.96 & 0.57 \\
    orchestration & 196 & 0.86 & \textbf{0.95} & 0.91 & 0.94 \\
    tool-policy   & 302 & \textbf{0.88} & 0.81 & 0.81 & 0.62 \\
    \midrule
    \textbf{overall} & 2{,}057 & \textbf{0.88} & 0.86 & 0.82 & 0.71 \\
    \bottomrule
  \end{tabular}
\end{table}

\begin{figure}[t]
  \centering
  \includegraphics[width=0.78\linewidth]{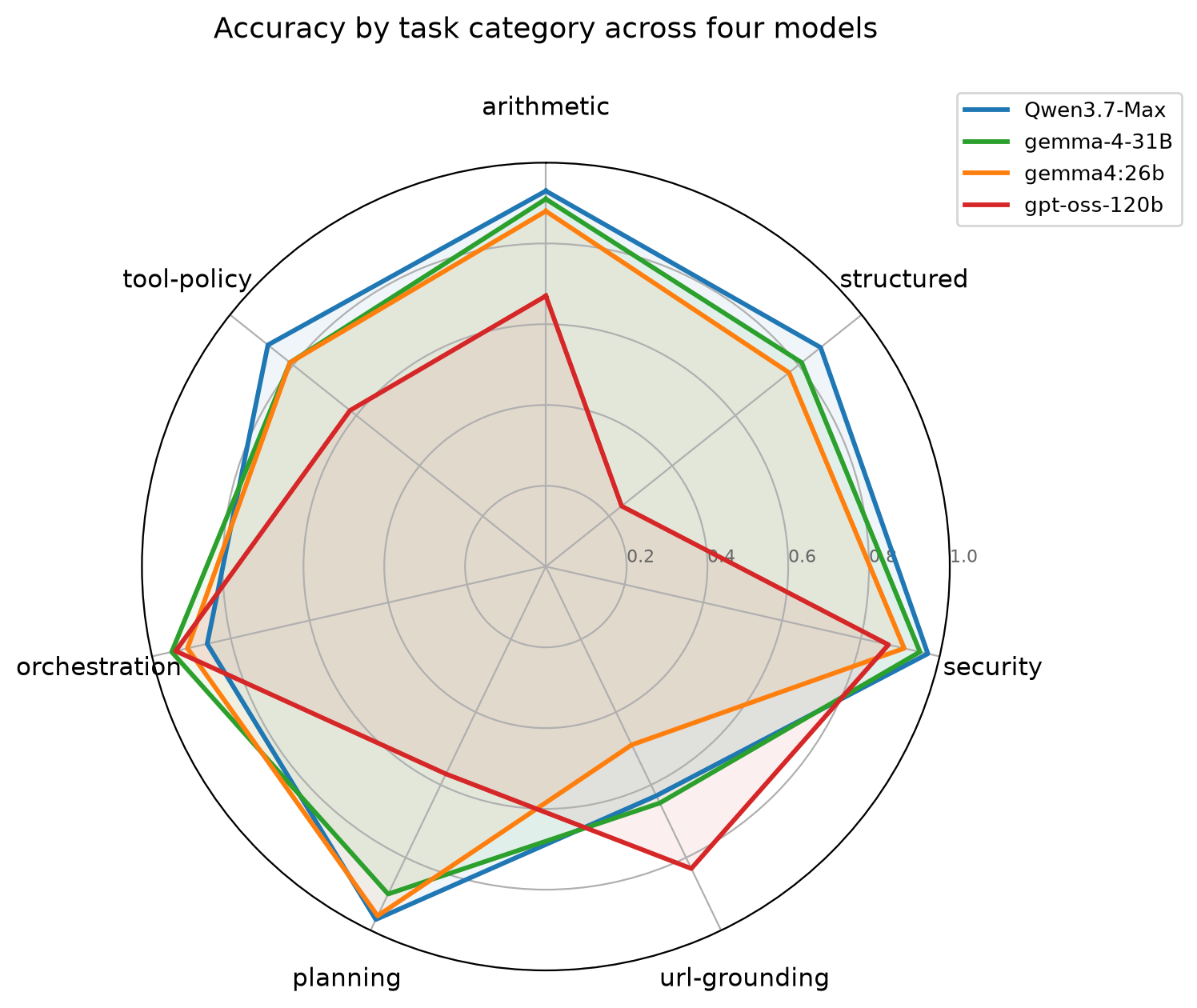}
  \caption{Zero-shot accuracy by task category for the four models---one axis per
  category (the task-weighted mean over its families of Table~\ref{tab:models_cat};
  radius $=$ success rate on $[0,1]$). The polygons expose each model's
  \emph{profile}: \texttt{gpt-oss-120b} (red) caves in on the \textsf{structured}
  ($0.24$), \textsf{planning} ($0.57$), and \textsf{tool-policy} ($0.62$) categories
  yet reaches farthest on \textsf{url-grounding} ($0.83$), while the other three trace
  a broadly similar high envelope---a shape an overall score cannot convey.}
  \label{fig:model_radar}
\end{figure}

\begin{table}[t]
  \caption{Operational cost of the same runs, measured by the platform: LLM calls and
  token usage per model over the full $2{,}057$-task suite (input\,+\,output tokens
  in millions). Read alongside Table~\ref{tab:models}, this shows capability and cost
  from one instrumented stream: \texttt{gemma-4-31B} and \texttt{gpt-oss-120b} spend
  the most tokens without leading on accuracy, while \texttt{gemma4:26b} is the
  leanest.}
  \label{tab:models_ops}
  \centering
  \begin{tabular}{lrrrr}
    \toprule
    Model & LLM calls & input (M) & output (M) & total (M) \\
    \midrule
    Qwen3.7-Max  & \textbf{6{,}965} & \textbf{3.86} & 1.96 & 5.83 \\
    gemma-4-31B  & 7{,}715 & 8.32 & 1.80 & 10.12 \\
    gemma4:26b   & 7{,}258 & 4.61 & \textbf{0.82} & \textbf{5.43} \\
    gpt-oss-120b & 7{,}158 & 6.56 & 2.25 & 8.81 \\
    \bottomrule
  \end{tabular}
\end{table}

\paragraph{Findings.}
Table~\ref{tab:models} gives each model a distinct \textbf{capability} profile
(Figure~\ref{fig:model_radar}), and the story is in the per-family spread rather than
the aggregate. Overall the four models rank \texttt{Qwen3.7-Max} ($0.88$) $>$
\texttt{gemma-4-31B} ($0.86$) $>$ \texttt{gemma4:26b} ($0.82$) $>$
\texttt{gpt-oss-120b} ($0.71$), but the top two are within two points and no model
dominates everywhere. \texttt{gpt-oss-120b} is a telling outlier that \emph{inverts}
the usual ordering: it is the best of the four on both grounding families
(\texttt{ug\_fetch} $0.78$, where every other model sits at $0.43$--$0.53$;
\texttt{ug\_imposter} $0.89$) yet by far the worst on the structured and multi-step
families (\texttt{sql\_analytics} $0.24$, \texttt{tool\_error\_recovery} $0.25$,
\texttt{multi\_tool\_planning} $0.44$)---a profile an overall score would hide
entirely. The families that discriminate most widely are exactly these tool/structured
tasks: \texttt{sql\_analytics} spans $0.24$--$0.87$, \texttt{tool\_error\_recovery}
$0.25$--$0.76$, and \texttt{multi\_tool\_planning} $0.44$--$0.96$, so a single family
separates the field by more than sixty points. \texttt{gemma-4-31B} leads on
\texttt{orchestrator} ($0.95$) and ties for the top on the security and
dependency-planning families, while the small local \texttt{gemma4:26b} is competitive
throughout ($0.82$ overall) and even leads \texttt{ar\_math} ($0.91$) and
\texttt{multi\_tool\_planning} ($0.96$). A few families are saturated across all four
(\texttt{sec\_sqli} $0.94$--$0.99$) and serve as sanity checks rather than ranking
signal. The platform captures the \textbf{operational} side of the same runs
(Table~\ref{tab:models_ops}): every model issues $7$k--$8$k LLM calls over the suite,
but token usage varies almost $2\times$---\texttt{gemma-4-31B} spends the most
($10.1$M) and \texttt{gemma4:26b} the least ($5.4$M), and \texttt{gpt-oss-120b} burns
$8.8$M for the lowest accuracy. Capability and cost are thus read from one instrumented
stream rather than separate experiments.

%% ============================================================================
\section{The BekchiAI-Platform}
\label{sec:platform}
%% ============================================================================
BekchiAI-Platform is a drop-in
observability and control layer. Enabling telemetry points a lightweight SDK at the
platform; the agent's task logic is unchanged. This is what lets us read capability and
cost (Section~\ref{sec:results}) from the same run, and it adds security observability
and a \textbf{remote kill-switch} on top.

\paragraph{Architecture and event model.}
The platform has four parts: a \textbf{client SDK} that emits telemetry and polls for
control commands; an \textbf{ingestion + API} and REST endpoints for agents, sessions, traces, and findings;
a store with a wide \texttt{events} table (JSONB payloads plus relational metadata); and a \textbf{React dashboard}.
Ingestion is idempotent on a event id, authenticated with per-project API keys,
and multi-tenant. Agent activity is modelled as \emph{sessions} of linked events
forming a trace tree. Core types are \texttt{session\_start}; \texttt{llm\_call}
(model, input/output tokens, latency); \texttt{tool\_call} / \texttt{tool\_result}
(name, arguments, success); \texttt{url\_access} (host, status, allowlisted);
\texttt{plan} (intended tools and URLs); \texttt{denied} (policy events); and
\texttt{outcome}. Table~\ref{tab:map} shows how each benchmark behavior maps onto these
events, so no separate evaluation harness is needed.

\begin{table}[t]
  \caption{How each benchmark behavior maps onto BekchiAI-Platform signals. One
  instrumented run yields all four.}
  \label{tab:map}
  \centering
  \begin{tabular}{lll}
    \toprule
    Model behaviour & Platform signal (event) & Reveals \\
    \midrule
    task correctness      & \texttt{outcome} (success/mismatch)          & accuracy, per family \\
    model call            & \texttt{llm\_call} (input/output tokens, latency) & token usage and speed \\
    tool use              & \texttt{tool\_call}, plan-vs-actual          & tool competence, adherence \\
    URL arguments         & \texttt{url\_access} (grounded/hallucinated) & URL-hallucination, source-match \\
    \bottomrule
  \end{tabular}
\end{table}

\paragraph{Behavioral-security analytics.}
Over the event store the platform computes two signals.
\emph{Argument grounding}: from each session's \texttt{plan} and allowlisted accesses
it builds a trusted-host set and labels each accessed host grounded, ungrounded, or
\texttt{imposter\_url} (an ungrounded host within Levenshtein distance $\le 2$ of a
trusted one---a likely typosquat), reporting the grounding rate. 

\emph{Plan-vs-actual
consistency}: it compares the intended tool set to the executed \texttt{tool\_call}
sequence, yielding intersection / unplanned / missing sets and a per-session verdict.
These are the platform-side counterparts of the benchmark's grounding and
tool-adherence metrics.

\paragraph{Policy enforcement and remote control.}
Operators define \texttt{allow}/\texttt{block}/\texttt{alert} policies over host globs
and tool names; at ingest, \texttt{block} matches emit a synthetic \texttt{denied}
event, giving a durable audit of prevented actions. The platform is also bidirectional:
an operator issues \texttt{stop}/\texttt{pause}/\texttt{resume}/\texttt{revoke} from the
dashboard, the agent polls the control endpoint, applies pending commands, and
acknowledges; commands move \texttt{pending}$\to$\texttt{delivered}$\to$\texttt{applied}
with timestamps. In the benchmark run this let us abort a running agent remotely within
a single poll cycle. Enforcement is cooperative: an agent that cannot reach the channel
is not controllable.

%% ============================================================================
\section{Conclusion}
\label{sec:conclusion}
%% ============================================================================
We introduced the BekchiAI-Benchmark, a suite of $13$ tool-using ReAct agents across
seven categories ($2{,}057$ committed tasks) that measures the \emph{agentic skills} of
LLMs---tool selection and sequencing, planning under dependencies, security judgment,
and argument grounding---rather than a single blended accuracy. Three design choices
keep the suite honest: gold is computed by a verifier (canonical SQL, the exact
schedule of a dependency DAG, or a closed-form lambda) and so cannot drift; the
discriminating cases are hand-authored one by one rather than templated; and the
security families replace the usual oracle detector with a deliberately imperfect
signature scanner, so a score reflects the model's own judgment rather than
tool-copying. We pair accuracy with behavioral metrics---tool-call adherence,
plan-vs-actual consistency, URL grounding, and per-model token cost---all read from one
instrumented stream via the drop-in BekchiAI-Platform.

Across four models, the results argue for reporting the \emph{per-family spread} rather
than an aggregate. The top two models finish within two points overall, yet no model
dominates every skill: \texttt{gpt-oss-120b} leads both grounding families while
collapsing on the structured and multi-step tool tasks (e.g.\ \texttt{sql\_analytics}
$0.24$ vs.\ $0.87$)---an inversion a single leaderboard number erases. The tool and
structured families discriminate most sharply, separating the field by more than sixty
points. Because the
platform records token cost on the same runs, capability and efficiency can be weighed
together---and the most accurate model is not the cheapest. As agents take on real,
consequential actions, we argue that skill- and cost-resolved, hard-to-game evaluation
of this kind is a prerequisite for trustworthy deployment.

The benchmark is deliberately compact and evaluates one task per rollout; natural
extensions include broader model coverage, additional adversarial rounds and skill
categories, longer-horizon and multi-turn interactions, and applying the platform's
grounding and plan-vs-actual analytics to live agent deployments. We release the
benchmark, tool library, metric scripts, and we released to platform on web and mobile markets to support this work.

\paragraph{Reproducibility.}
The benchmark---provided test sets, the tool library, harness,
and metrics---is released at
\url{https://github.com/bekchiai/bekchiai-benchmark}.

\bibliographystyle{unsrtnat}
\bibliography{references}

\end{document}